# Computing Presuppositions by Contextual Reasoning

**Christof Monz**
Institute for Logic, Language and Computation (ILLC)
University of Amsterdam
Plantage Muidergracht 24, 1018 TV Amsterdam
The Netherlands
E-mail:

**Abstract**

This paper describes how automated deduction methods for natural language processing can be applied more efficiently by encoding context in a more elaborate way. Our work is based on formal approaches to context, and we provide a tableau calculus for contextual reasoning. This is explained by considering an example from the problem area of presupposition projection.

## Introduction

The notion of context plays an important role in formal theories of natural language processing (NLP), one of the major subareas of Artificial Intelligence. Several definitions of context have been used in computational linguistics, depending on the application. Amongst others, context has been used for resolving lexical ambiguity (Buvač 1996), pronoun resolution (Monz & de Rijke 1999), and presupposition projection (Karttunen 1974). In this paper, we focus on the last application.

Before we go into detail, we give a rough definition of presupposition. We say that an utterance $\varphi$ presupposes a fact $\pi$, if uttering $\varphi$ only makes sense if the context (e.g., world knowledge or earlier utterances in the same conversation) provides enough information to conclude that $\pi$ is the case. Consider an example.

(1)a. Sue's husband is out of town.

b. If Sue is married, then her husband is out of town.

Presupposition triggers like *Sue's husband* are resolved against their local context. If this context provides the presupposed information, then the presupposition does not project. If, on the other hand, the local context does not provide the presupposed information, it does project.

As to the explicit representation of context itself, we try to keep things as simple as possible and identify a context with a set of first-order formulas.

There are many formal theories in Linguistics that describe presupposition projection by using context, e.g. (Karttunen 1974), but only a few NLP systems computing presuppositions have been actually implemented, see, e.g., DORIS (Blackburn *et al.* 1999). To compute the presuppositions which project, it is necessary to employ automated reasoning techniques, where, as we said before, a presupposition $\pi$ does not project if it can be deduced from its local context. Computing presupposition projection involves both concepts: *context* and *reasoning*.

If a semantic representation of a natural language discourse contains the presuppositions $\pi_1, \ldots, \pi_n$, where $\Gamma_1, \ldots, \Gamma_n$ are the corresponding local contexts (sets of first-order formulas), then the most straightforward way to compute the projected presuppositions is by having run a theorem prover on the queries (or projection problems) $\Gamma_1 \vdash \pi_1, \ldots, \Gamma_n \vdash \pi_n$. Its major shortcoming is that deductions are carried out independently of each other. But, often it is the case that the contexts $\Gamma_1, \ldots, \Gamma_n$ share formulas. Then, some deduction steps are carried out several times, causing a decrease of efficiency.

To avoid redundant treatment of subcontexts, we need a richer language that enables us to express nesting of contexts. Here, we use the `in`-predicate, cf. (Attardi & Simi 1994a; 1994b), which takes two arguments. The first argument is a set of formulas and the second is a formula. $\text{in}(\Gamma, \varphi)$ is true if $\Gamma \vdash \varphi$.[1] Since $\varphi$ itself can contain an `in`-predicate, we are able to nest contexts. E.g., $\text{in}(\Gamma, \varphi \wedge \text{in}(\Delta, \psi))$ is true if $\Gamma \vdash \varphi$ and $\Gamma \cup \Delta \vdash \psi$. In this case $\Delta$ is a local context for $\psi$.

A language containing the `in`-predicate functions like a meta-language of reasoning. In the sequel, we give a tableau calculus for such a language.

## Presupposition and Context

A particular instance of the presupposition projection problem was given in (1). Here, the noun phrase *Sue's husband* carries the presupposition *Sue is married*. In general, following (Blamey 1986), we use an additional connective $\pi/\varphi$, meaning that $\varphi$ presupposes $\pi$.

**Definition 1 (The Language $\mathcal{L}^{pre}$)** To indicate the presuppositions of a formula, the language of first-order predicate logic $\mathcal{L}$ is augmented with a binary presupposition connective '/'. Its left argument is a first-order formula and its right argument a formula of $\mathcal{L}^{pre}$.

$$\varphi \quad ::= \quad R(t_1 \ldots t_n) \mid \neg \varphi \mid \varphi \wedge \psi \mid \varphi \rightarrow \psi \mid \varphi \vee \psi$$
$$\mid \forall x \varphi \mid \exists x \varphi \mid \pi/\varphi$$

---

[1] We slightly diverge from (Attardi & Simi 1994a; 1994b) where the argument positions of `in` are interchanged, i.e., $\text{in}(\varphi, \Gamma)$ means that $\varphi$ can be deduced from $\Gamma$.

where $\pi \in \mathcal{L}$.

The set of contexts $\mathcal{C}$ is identified with the finite subsets of $\mathcal{L}$.

Above, we sketched how the projected presupposition of (1) can be computed. (Karttunen 1974) defines a function pres accomplishing this task.

**Definition 2 (Presupposition)** The function pres is defined recursively. It takes two arguments: A formula of $\mathcal{L}^{pre}$ and a context $\Gamma \in \mathcal{C}$, pres : $\mathcal{L}^{pre} \times \mathcal{C} \to POW(\mathcal{L})$.

$$
\begin{aligned}
&\text{(i)} & \text{pres}(\varphi)^\Gamma &= \emptyset \text{ if } \varphi \text{ is atomic} \\
&\text{(ii)} & \text{pres}(\neg\varphi)^\Gamma &= \text{pres}(\varphi)^\Gamma \\
&\text{(iii)} & \text{pres}(\varphi \wedge \psi)^\Gamma &= \text{pres}(\varphi)^\Gamma \cup \text{pres}(\psi)^{\Gamma \cup \{\text{as}(\varphi)\}} \\
&\text{(iv)} & \text{pres}(\varphi \to \psi)^\Gamma &= \text{pres}(\varphi)^\Gamma \cup \text{pres}(\psi)^{\Gamma \cup \{\text{as}(\varphi)\}} \\
&\text{(v)} & \text{pres}(\varphi \vee \psi)^\Gamma &= \text{pres}(\varphi)^\Gamma \cup \text{pres}(\psi)^{\Gamma \cup \{\text{as}(\neg\varphi)\}} \\
&\text{(vi)} & \text{pres}(\forall x \varphi)^\Gamma &= \text{pres}(\varphi)^\Gamma \\
&\text{(vii)} & \text{pres}(\exists x \varphi)^\Gamma &= \text{pres}(\varphi)^\Gamma \\
&\text{(viii)} & \text{pres}(\pi/\varphi)^\Gamma &= \begin{cases} \text{pres}(\varphi)^\Gamma \cup \{\pi\} & \text{if } \Gamma \not\vdash \pi \\ \text{pres}(\varphi)^\Gamma & \text{if } \Gamma \vdash \pi \end{cases}
\end{aligned}
$$

The last rule is the one we want to focus on this paper, because it involves contextual reasoning.

The function as returns the assertive content of a formula of $\mathcal{L}^{pre}$, cf. (Karttunen & Peters 1979). Roughly speaking, the assertive part of a formula $\varphi \in \mathcal{L}^{pre}$ is obtained by substituting all subformulas of the form $\pi/\psi$ by $\psi$. The assertive part can be recursively computed by applying the function as to $\varphi$.

**Definition 3 (Assertive Content)** as is a function from $\mathcal{L}^{inc}$ to $\mathcal{L}$:

$$
\begin{aligned}
\text{as}(R(t_1 \ldots t_n)) &= R(t_1 \ldots t_n) \\
\text{as}(\neg\varphi) &= \neg\text{as}(\varphi) \\
\text{as}(\varphi \bullet \psi) &= \text{as}(\varphi) \bullet \text{as}(\psi), \text{ where } \bullet \in \{\wedge, \to, \vee\} \\
\text{as}(\pi/\varphi) &= \text{as}(\varphi) \\
\text{as}(Qx\varphi) &= Qx\text{as}(\varphi), \text{ where } Q \in \{\forall, \exists\}
\end{aligned}
$$

Let us consider an example. (2) displays an admittedly quite artificial discourse, but we hope that it illustrates the point.

(2)a. John is married. ($\varphi$)

    b. If <u>John's wife</u> has a brother, then John hasn't met <u>his brother-in-law</u>, yet. $(\varphi/\psi \to (\varphi \wedge \psi)/\chi)$

    c. <u>John's wife</u> has quite a few relatives. $(\varphi/\omega)$

The semantic representations of the three sentences are displayed in parentheses after each sentence and the presupposition triggers are underlined.

Applying pres to the whole discourse (the conjunction of (2.a), (2.b), and (2.c)), it correctly computes that no presupposition projects:

$$\text{pres}(\varphi \wedge ((\varphi/\psi \to (\varphi \wedge \psi)/\chi) \wedge \varphi/\omega))^\emptyset = \emptyset$$

To see why this is the case, let us have a closer look at the way pres recursively computes the presuppositions. The contextual parameter of pres is initialized to $\emptyset$, as there is no further context preceding (2).

$$
\begin{aligned}
&\text{pres}(\varphi \wedge ((\varphi/\psi \to (\varphi \wedge \psi)/\chi) \wedge \varphi/\omega))^\emptyset \\
&= \text{pres}(\varphi)^\emptyset \cup \text{pres}((\varphi/\psi \to (\varphi \wedge \psi)/\chi) \wedge \varphi/\omega)^{\{\varphi\}} & \text{by (iii)} \\
&= \text{pres}((\varphi/\psi \to (\varphi \wedge \psi)/\chi) \wedge \varphi/\omega)^{\{\varphi\}} & \text{by (i)} \\
&= \text{pres}((\varphi/\psi \to (\varphi \wedge \psi)/\chi))^{\{\varphi\}} \cup \text{pres}(\varphi/\omega)^{\{\varphi,\psi\}} & \text{by (iii)} \\
&= \text{pres}(\varphi/\psi)^{\{\varphi\}} \cup \text{pres}((\varphi \wedge \psi)/\chi))^{\{\varphi,\psi\}} \\
&\quad \cup \text{pres}(\varphi/\omega)^{\{\varphi,\psi \to \chi\}} & \text{by (iv)} \\
&= \text{pres}((\varphi \wedge \psi)/\chi))^{\{\varphi,\psi\}} \cup \text{pres}(\varphi/\omega)^{\{\varphi,\psi \to \chi\}} & \text{by (viii)} \\
&\quad \text{where pres}(\varphi/\psi)^{\{\varphi\}} = \emptyset \text{ because } \{\varphi\} \vdash \varphi \\
&= \text{pres}(\varphi/\omega)^{\{\varphi,\psi \to \chi\}} & \text{by (viii)} \\
&\quad \text{where pres}((\varphi \wedge \psi)/\chi)^{\{\varphi,\psi\}} = \emptyset \text{ because } \{\varphi,\psi\} \vdash \varphi \wedge \psi \\
&= \emptyset & \text{by (viii)} \\
&\quad \text{where pres}(\varphi/\omega)^{\{\varphi,\psi \to \chi\}} = \emptyset \text{ because } \{\varphi,\psi \to \chi\} \vdash \varphi
\end{aligned}
$$

The application of pres to (2) involves three inferences: $\{\varphi\} \vdash \varphi$, $\{\varphi,\psi\} \vdash \varphi \wedge \psi$, and $\{\varphi,\psi \to \chi\} \vdash \varphi$. Here, the premise $\varphi$ is used three times and the proving method has to apply the same set of rules three times to the same formula. As we mentioned earlier, this is due to the fact that the contexts (premises) are considered to be independent of each other. From a computational point of view, this redundancy is rather inefficient. Of course, the example is very simple, but in general, $\varphi$ could have been arbitrarily complex. In (2), the sentences containing presuppositions, (2.b) and (2.c), occur in a context consisting of one sentence, viz. (2.a), but it could have been a much larger context.

This reasoning task can be done more efficiently if we take the flow of contextual information into account; i.e., the way contexts are nested. To express nesting of contexts we use a language containing the in-predicate:

**Definition 4 (The Language $\mathcal{L}^{con}$)** $\mathcal{L}^{con}$ is defined recursively as follows, where $\Gamma \in \mathcal{C}$:

$$\varphi ::= R(t_1 \ldots t_n) \mid \neg\varphi \mid \varphi \wedge \psi \mid \varphi \to \psi \mid \varphi \vee \psi$$
$$\mid \forall x \varphi \mid \exists x \varphi \mid \text{in}(\Gamma,\varphi) \mid \top$$

We do not use $\mathcal{L}^{con}$ to express the semantics of a natural language discourse, but only for expressing which presupposition triggers have to be evaluated against which context. The purpose of $\mathcal{L}^{con}$ is to express these projection problems in a non-redundant fashion.

Let us reconsider example (2). The antecedent of (2.b) contains the presupposition, that John is married, formalized as $\varphi$. The context of this presupposition only consists of (2.a), also formalized as $\varphi$. This projection problem can be stated in $\mathcal{L}^{con}$ as $\text{in}(\{\varphi\},\varphi)$. The presupposition of the succedent of (2.b), i.e., John is married and that his wife has a brother, formalized as $\varphi \wedge \psi$ is evaluated against (2.a) and the antecedent of (2.b): $\text{in}(\{\varphi,\psi\},\varphi \wedge \psi)$. The two projection problems can be compactly expressed by a single formula of $\mathcal{L}^{con}$: $\text{in}(\{\varphi\},\varphi \wedge \text{in}(\{\psi\},\varphi \wedge \psi))$. The second in-predicate is nested under a context where $\varphi$ already holds, and it is not necessary to express again that $\varphi$ holds. The last sentence, (2.c), again presupposes $\varphi$. Its context is (2.a) and(2.b), therefore, $\text{in}(\{\varphi,\psi \to \chi\},\varphi)$ has to hold. Putting the three projection problems together, we get

(3) $\text{in}(\{\varphi\},\varphi \wedge \text{in}(\{\psi\},\varphi \wedge \psi) \wedge \text{in}(\{\psi \to \chi\},\varphi))$

Note, that we did not embed the third in-predicate in the second in-predicate, yielding

(4) $\text{in}(\{\varphi\}, \varphi \land \text{in}(\{\psi\}, \varphi \land \psi \land \text{in}(\{\psi \to \chi\}, \varphi)))$

This is not possible because $\psi$ is a local assumption (context) only accessible within the conditional.

The obvious question is how we get from a semantic representation in $\mathcal{L}^{pre}$ to a description of the projection problems in $\mathcal{L}^{con}$. To this end, we define a translation function $\tau$ which takes two arguments: a formula of $\mathcal{L}^{pre}$ and a context. Before we can give the formal definition of $\tau$, we have to define how the potential presuppositions of a formula can be computed.

**Definition 5 (Potential Presupposition)** The potential presupposition of a formula $\varphi \in \mathcal{L}^{inc}$ are all formulas $\pi \in \mathcal{L}$ that occur as subformulas of $\varphi$ of the form $\pi/\psi$.

$$\begin{aligned}
\text{pp}(R(t_1 \ldots t_n)) &= \emptyset \\
\text{pp}(\neg \varphi) &= \text{pp}(\varphi) \\
\text{pp}(\varphi \bullet \psi) &= \text{pp}(\varphi) \cup \text{pp}(\psi) \text{ if } \bullet \in \{\land, \to, \lor\} \\
\text{pp}(\pi/\varphi) &= \text{pp}(\varphi) \cup \{\pi\} \\
\text{pp}(Qx\varphi) &= \text{pp}(\varphi) \text{ if } Q \in \{\forall, \exists\}
\end{aligned}$$

pp simply collects all presuppositions without checking whether they are entailed by there local context.

Now, we can define a translation $\tau$ from $\mathcal{L}^{pre}$ to $\mathcal{L}^{con}$. The function $\tau$ is defined in Table 1. Since it is rather complex and we have only limited space, we just try to sketch its rationale. $\tau$ is defined recursively, taking two parameters: a formula of $\mathcal{L}^{pre}$ and a context $\Gamma$.

| $\tau : \mathcal{L}^{pre} \times \mathcal{C} \to \mathcal{L}^{con}$ | |
|---|---|
| $R(t_1 \ldots t_n)^{\tau, \Gamma} = \top$ | (1) |
| $(\neg \varphi)^{\tau, \Gamma} = \begin{cases} (\varphi)^{\tau, \Gamma} & \text{if } \text{pp}(\varphi) \neq \emptyset \\ \top & \text{if } \text{pp}(\varphi) = \emptyset \end{cases}$ | (2a) (2b) |
| $(\varphi \bullet \psi)^{\tau, \Gamma}$ $= \begin{cases} \text{in}(\Gamma, (\varphi)^{\tau, \emptyset} \land (\psi)^{\tau, \{\text{as}(\varphi)\}}) & \text{if } \text{pp}(\varphi) \neq \emptyset, \Gamma \neq \emptyset \\ (\varphi)^{\tau, \emptyset} \land (\psi)^{\tau, \{\text{as}(\varphi)\}} & \text{if } \text{pp}(\varphi) \neq \emptyset, \Gamma = \emptyset \\ (\psi)^{\tau, \Gamma \cup \{\varphi\}} & \text{if } \text{pp}(\varphi) = \emptyset \end{cases}$ where $\bullet \in \{\land, \to\}$ | (3a) (3b) (3c) |
| $(\varphi \lor \psi)^{\tau, \Gamma}$ $= \begin{cases} \text{in}(\Gamma, (\varphi)^{\tau, \emptyset} \land (\psi)^{\tau, \{\text{as}(\neg \varphi)\}}) & \text{if } \text{pp}(\varphi) \neq \emptyset, \Gamma \neq \emptyset \\ (\varphi)^{\tau, \emptyset} \land (\psi)^{\tau, \{\text{as}(\neg \varphi)\}} & \text{if } \text{pp}(\varphi) \neq \emptyset, \Gamma = \emptyset \\ (\psi)^{\tau, \Gamma \cup \{\neg \varphi\}} & \text{if } \text{pp}(\varphi) = \emptyset \end{cases}$ | (4a) (4b) (4c) |
| $(\pi/\varphi)^{\tau, \Gamma} = \begin{cases} \text{in}(\Gamma, \pi \land (\varphi)^{\tau, \emptyset}) & \text{if } \text{pp}(\varphi) \neq \emptyset, \Gamma \neq \emptyset \\ \pi \land (\varphi)^{\tau, \emptyset} & \text{if } \text{pp}(\varphi) \neq \emptyset, \Gamma = \emptyset \\ \text{in}(\Gamma, \pi) & \text{if } \text{pp}(\varphi) = \emptyset, \Gamma \neq \emptyset \\ \pi & \text{if } \text{pp}(\varphi) = \emptyset, \Gamma = \emptyset \end{cases}$ | (5a) (5b) (5c) (5d) |
| $(Qx\varphi)^{\tau, \Gamma} = \begin{cases} \text{in}(\Gamma, Qx(\varphi)^{\tau, \emptyset}) & \text{if } \text{pp}(\varphi) \neq \emptyset, \Gamma \neq \emptyset \\ Qx(\varphi)^{\tau, \Gamma} & \text{if } \text{pp}(\varphi) \neq \emptyset, \Gamma = \emptyset \\ \top & \text{if } \text{pp}(\varphi) = \emptyset \end{cases}$ where $Q \in \{\forall, \exists\}$ | (6a) (6b) (6c) |

Table 1: Translating from $\mathcal{L}^{pre}$ to $\mathcal{L}^{con}$

Let $\chi$ be in $\mathcal{L}^{pre}$. If the main operator is unary and $\chi$ contains potential presuppositions, then we proceed with translating the immediate subformula of $\chi$. If $\chi$ is of the form $\varphi \land \psi$, we check whether $\varphi$ contains potential presuppositions. If this is the case, $\chi$ translates as $\text{in}(\Gamma, (\varphi)^{\tau, \emptyset} \land (\psi)^{\tau, \{\text{as}(\varphi)\}})$. The first argument of in is the context and $(\varphi)^{\tau, \emptyset}$ gets $\emptyset$ as the context, because the translation of $\varphi$ will be embedded in $\Gamma$ by the in-predicate. Similarly for $(\psi)^{\tau, \{\text{as}(\varphi)\}}$, but here the context is augmented by the assertive content of $\varphi$, which was not available for $\varphi$ itself. No in-predicate is introduced if the context parameter is empty. Note that the way how contexts are augmented, see, for instance, $(\psi)^{\tau, \{\text{as}(\varphi)\}}$, follows the definition of pres.

Reconsidering (2), the translation of its semantic representation in $\mathcal{L}^{pre}$

$$\varphi \land ((\varphi/\psi \to (\varphi \land \psi)/\chi) \land \varphi/\omega)$$

proceeds as follows:

$(\varphi \land ((\varphi/\psi \to (\varphi \land \psi)/\chi) \land \varphi/\omega))^{\tau, \emptyset}$
by (3c):
$= ((\varphi/\psi \to (\varphi \land \psi)/\chi) \land \varphi/\omega)^{\tau, \{\varphi\}}$
by (3a):
$= \text{in}(\{\varphi\}, (\varphi/\psi \to (\varphi \land \psi)/\chi)^{\tau, \emptyset} \land (\varphi/\omega)^{\tau, \{\psi \to \chi\}})$
by (3b):
$= \text{in}(\{\varphi\}, (\varphi/\psi)^{\tau, \emptyset} \land ((\varphi \land \psi)/\chi)^{\tau, \{\chi\}} \land (\varphi/\omega)^{\tau, \{\psi \to \chi\}})$
by (5d):
$= \text{in}(\{\varphi\}, \varphi \land ((\varphi \land \psi)/\chi)^{\tau, \{\chi\}} \land (\varphi/\omega)^{\tau, \{\psi \to \chi\}})$
by (5c):
$= \text{in}(\{\varphi\}, \varphi \land \text{in}(\{\chi\}, \varphi \land \psi) \land (\varphi/\omega)^{\tau, \{\psi \to \chi\}})$
by (5c):
$= \text{in}(\{\varphi\}, \varphi \land \text{in}(\{\chi\}, \varphi \land \psi) \land \text{in}(\{\psi \to \chi\}, \varphi))$

Now we have defined an algorithmic way to state a presupposition projection problem in a compact way by translating it to $\mathcal{L}^{con}$. The main advantage of expressing projection problems in $\mathcal{L}^{con}$ is that it allows for a more efficient way of reasoning; see below.

## Contextual Reasoning

In the previous section, we have seen how translating from $\mathcal{L}^{pro}$ to $\mathcal{L}^{con}$ can help stating presupposition projection problems in a compact way. In this section, we provide a tableau calculus for the language $\mathcal{L}^{con}$. If $\varphi \in \mathcal{L}^{con}$ is valid, then all presuppositions are entailed by their local context and none of them projects.

The most important rule of our tableau calculus $\mathcal{T}^{con}$ is the rule ($\neg\text{in}$). Before we introduce the other rules, it is helpful to have a closer look at ($\neg\text{in}$), in order to understand the way context is represented in $\mathcal{T}^{con}$.

$$\frac{(i, \sigma) : \neg\text{in}(\{\varphi_1, \ldots, \varphi_n\}, \psi)}{\begin{array}{c}(j, \sigma \cup \{i\}) : \varphi_1 \\ \vdots \\ (j, \sigma \cup \{i\}) : \varphi_n \\ (j, \sigma \cup \{i\}) : \neg\psi\end{array}} (\neg\text{in})$$

To keep track of the contextual information, labels are attached to the nodes in the tableau. A label has two arguments. Its first argument $i$ is a natural number ($i \in \mathbb{N}$), which is the identifier of the context. I.e., if two nodes have the same number as the first argument of their labels, then they belong to the same context.

The second argument $\sigma$ is a set of natural numbers. This set contains the identifiers of the contexts that are accessible. We say that a context $\Gamma$ is accessible from a formula $\psi$, if there is a formula of the form $\text{in}(\Gamma, \varphi)$ and $\psi$ is a subformula of $\varphi$. For instance, considering the formula $\text{in}(\Gamma, \varphi \wedge \text{in}(\Delta, \psi))$, $\Gamma$ is accessible from $\varphi$ and $\text{in}(\Delta, \psi)$. Also $\Delta$ is accessible from $\psi$. Since accessibility is transitive, it holds that $\Gamma$ is accessible from $\psi$; but $\Delta$ is not accessible from $\varphi$.

The ($\neg\text{in}$)-rule is similar to the upwards direction (entering a context) of the (CS)-rule in (Buvač & Mason 1993):

$$\frac{\vdash_{\bar{\kappa}*\kappa_1} \varphi}{\vdash_{\bar{\kappa}} \text{ist}(\kappa_1, \varphi)} \text{ (CS)}$$

$\bar{\kappa}$ represents a sequence of contexts and the upwards direction of the rule says that if it is true in the context $\bar{\kappa}$ that $\varphi$ holds in the extension with $\kappa_1$, then $\varphi$ holds in the context $\bar{\kappa} * \kappa_1$ itself.

Comparing (CS) to ($\neg\text{in}$), we can say that $\bar{\kappa}$ corresponds to $\sigma \cup \{i\}$ and $j$, the identifier of the context extension with $\{\varphi_1, \ldots, \varphi_n, \neg\psi\}$, corresponds to $\kappa_1$.

Table 2 gives the complete set of tableau rules. Besides the rules for the $\text{in}$-predicate, it contains the usual rule for the boolean connectives and quantifiers. $\ell$ is a meta-variable over labels of the form $(i, \sigma)$. As the expansion rules for the regular logical connectives do not change the contextual information, the label of the parent and the label(s) of the daughter(s) are identified.

The contextual information carried by the labels becomes important when we want to define the closure conditions of a branch.

**Definition 6 (Closure of a Branch)** A branch of a tableau tree is closed if it contains two nodes of the form $(i, \sigma) : \varphi$ and $(j, \sigma') : \neg\psi$ such that
(a)  $\varphi$ and $\psi$ are unifiable, and
(b)  (i) $i = j$ or (ii) $i \in \sigma'$ or (iii) $j \in \sigma$

(a) is the standard condition on branch closure. (b) considers three cases. If $i = j$, then both literals belong to the same context. If $i \in \sigma'$, then $\varphi$ belongs to an extension of $j$. (iii) is analogous to the previous one.

Again, we consider the projection problem of (2). Applying the tableau expansion rules to the negation of (3), we try to derive a contradiction (a closed tableau).

To verify the negation of the projection problem in the initial context 0, we create a new context 1, where the contextual assumptions of the $\text{in}$-predicate hold, but not the second argument. Both nodes, the node belonging to the contextual assumption of the $\text{in}$-predicate and the negation of its second argument, carry a 1 as its context identifier and have 0 (the context in which the $\text{in}$-predicate occurred) as an element of the set of accessible contexts. The remaining steps either follow the same pattern or are just regular boolean tableau expansion rules. The pairs of nodes that allow to close a branch are connected by a dashed line. $(1, \{0\}) : \varphi$ and $(2, \{1, 0\}) : \neg\varphi$ allow to close the second branch from the left, because the first node is accessible from the second, as $1 \in \{1, 0\}$,

$$\frac{(i, \sigma) : \text{in}(\{\varphi_1, \ldots, \varphi_n\}, \psi)}{(j, \sigma \cup \{i\}) : \bigvee_{i=1}^{n} \neg\varphi_i \mid (j, \sigma \cup \{i\}) : \psi} \text{ (in)}$$

$$\frac{(i, \sigma) : \neg\text{in}(\{\varphi_1, \ldots, \varphi_n\}, \psi)}{(j, \sigma \cup \{i\}) : \varphi_1} \text{ (}\neg\text{in)}$$
$$\vdots$$
$$(j, \sigma \cup \{i\}) : \varphi_n$$
$$(j, \sigma \cup \{i\}) : \neg\psi$$

$$\frac{\ell : \varphi_1 \wedge \varphi_2}{\ell : \varphi_1 \atop \ell : \varphi_2} \text{ (}\wedge\text{)} \qquad \frac{\ell : \neg(\varphi_1 \wedge \varphi_2)}{\ell : \neg\varphi_1 \vee \neg\varphi_2} \text{ (}\neg\wedge\text{)}$$

$$\frac{\ell : \varphi_1 \vee \varphi_2}{\ell : \varphi_1 \mid \ell : \varphi_2} \text{ (}\vee\text{)} \qquad \frac{\ell : \neg(\varphi_1 \vee \varphi_2)}{\ell : \neg\varphi_1 \wedge \neg\varphi_2} \text{ (}\neg\vee\text{)}$$

$$\frac{\ell : \varphi_1 \rightarrow \varphi_2}{\ell : \neg\varphi_1 \mid \ell : \varphi_2} \text{ (}\rightarrow\text{)} \qquad \frac{\ell : \neg(\varphi_1 \rightarrow \varphi_2)}{\ell : \varphi_1 \wedge \neg\varphi_2} \text{ (}\neg\rightarrow\text{)}$$

$$\frac{\ell : \neg\neg\varphi}{\ell : \varphi} \text{ (}\neg\neg\text{)}$$

$$\frac{\ell : \forall x \varphi}{\ell : \varphi[x/X]} \text{ (}\forall\text{)} \qquad \frac{\ell : \exists x \varphi}{\ell : \varphi[x/f(X_1 \ldots X_n)]} \text{ (}\exists\text{)}$$

$$\frac{\ell : \neg\forall x \varphi}{\ell : \exists x \neg\varphi} \text{ (}\neg\forall\text{)} \qquad \frac{\ell : \neg\exists x \varphi}{\ell : \forall x \neg\varphi} \text{ (}\neg\exists\text{)}$$

where for ($\text{in}$) and ($\neg\text{in}$): $j$ is a fresh natural number that does not occur elsewhere in the tableau, ($\forall$): $X$ is free in $\varphi$, and ($\exists$): $X_1 \ldots X_n$ are the free variables in $\varphi$.

Table 2: The set of tableaux rules

which is the set of contexts that are accessible from the second node. The first and the third branch can be closed because both nodes belong to the same context. The two rightmost branches can be closed, because 1 is accessible from 3.

## An Application to Dialogue Systems

Dialogue systems such as route planning systems, e.g., Trindi (Cooper et al. 1999), require that the system can deal with presuppositions. In dialogues, presuppositions play an important role because it is convenient to take some things for granted as they have been mentioned before or they are part of common knowledge. For instance, in the following dialogue, the presupposition trigger *my start*, assumes that it a start-location has been mentioned before.

**User:** I would like to go to Paris.

**System:** Where do you start?

**User:** My start is Amsterdam

**System:** When do you travel?   …

In order to detect whether the user presupposes material that has not been mentioned in the preceding discourse and which is not part of the system's knowledge base, it is necessary that the system can compute those presuppositions that project. If a presupposition projects, the system can get back to the user and ask him for futher explanations. This way of detecting missing information by using deductive methods is more elegant than simply looking whether a term has occurred before. The system behaves more intelligent and has to ask less questions about things which are obvious for the user as he has told the system before—although in other words. In order to improve the acceptability of dialogue system it is mandatory that the user can convey information in a convenient and non-redundant way without having to repeat things.

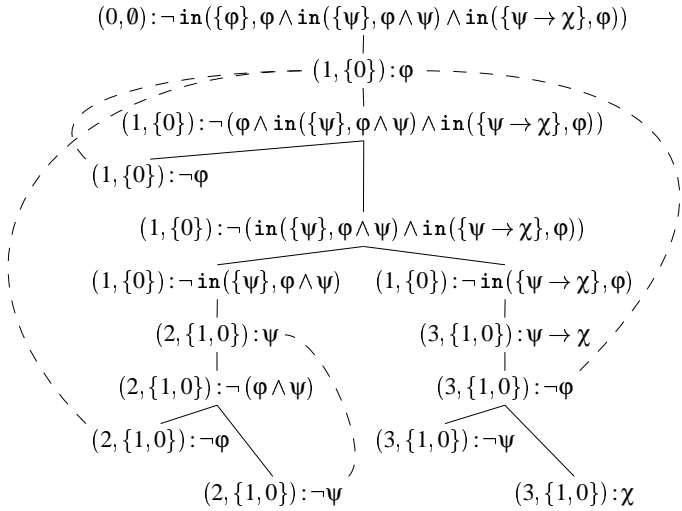

Figure 1: The tableau for the projection problem of (2)

Although the way context is represented in the Trindi system differs from our approach, it is possible to adapt our techniques to the Trindi system, cf. (Monz 1999).

## Conclusions and Future Work

Computing the presuppositions of a natural language discourse is an important task for a natural language processing system. Employing a language like $\mathcal{L}^{con}$ allows for a non-redundant way of stating presupposition problems. To this end, we gave a translation from $\mathcal{L}^{pre}$ to $\mathcal{L}^{con}$. In addition, a tableau calculus $\mathcal{T}^{con}$ has been presented, which allows to compute presupposition projection more efficiently than approaches considering the projection problems independent of each other.

In this paper, contexts were simply identified with sets of first-order formulas. Our future work will focus on a more complex representation of context, as it recently emerged in computational linguistics, cf. (van der Sandt 1992). His approach seems to be more appropriate for describing some projection phenomena that cannot be explained within a framework that uses a simple notion of context like we did. Nevertheless, we think that the translation function and the tableau calculus are defined generally enough to be adaptable to a more refined representation of context.

**Acknowledgments.** Thanks go to Marco Aiello and Maarten de Rijke for helpful comments. The author was supported by the Physical Sciences Council with financial support from the Netherlands Organization for Scientific Research (NWO), project 612-13-001.